\documentclass{article}

\usepackage{PRIMEarxiv}
\usepackage{float}
\restylefloat{table}
\usepackage[utf8]{inputenc} 
\usepackage[T1]{fontenc}    
\usepackage{hyperref}       
\usepackage{url}            
\usepackage{booktabs}       
\usepackage{amsfonts}       
\usepackage{nicefrac}       
\usepackage{microtype}      
\usepackage{lipsum}
\usepackage{fancyhdr}       
\usepackage{graphicx}       
\graphicspath{{media/}}     

\title{MTTN: Multi-Pair Text to Text Narratives for Prompt Generation
\thanks{\textit{\underline{Citation}}: 
\textbf{Ghosh et al. MTTN: Multi-Pair Text to Text Narratives for Prompt Generation.}}}

\author{
  Archan Ghosh, Debgandhar Ghosh, Madhurima Maji, Suchinta Chanda, Kalporup Goswami \\
  Kolkata, India\\
  \texttt{\{gharchan, debgandhar4000, rmadhurima99, suchintachanda, kalporup123\}@gmail.com} \\
}

\begin{document}
\maketitle

\begin{abstract}
The increased interest in diffusion models has opened up opportunities for advancements in generative text modeling. These models can produce impressive images when given a well-crafted prompt, but creating a powerful or meaningful prompt can be hit-or-miss. To address this, we have created a large-scale dataset that is derived and synthesized from real prompts and indexed with popular image-text datasets such as MS-COCO and Flickr. We have also implemented stages that gradually reduce context and increase complexity, which will further enhance the output due to the complex annotations created. The dataset, called MTTN, includes over 2.4 million sentences divided into 5 stages, resulting in a total of over 12 million pairs, and a vocabulary of over 300,000 unique words, providing ample variation. The original 2.4 million pairs are designed to reflect the way language is used on the internet globally, making the dataset more robust for any model trained on it.

\end{abstract}


\section{Introduction}
 We propose a new data set, MTTN(read mutton), for generating prompts that can be used in diffusion models\cite{DBLP:journals/corr/Sohl-DicksteinW15}\cite{DBLP:journals/corr/abs-2006-11239}\cite{DBLP:journals/corr/abs-1907-05600}. In this dataset, we present a collection of prompts for text-to-text generation tasks. Diffusion models\cite{DBLP:journals/corr/Sohl-DicksteinW15}\cite{DBLP:journals/corr/abs-2006-11239}\cite{DBLP:journals/corr/abs-1907-05600} are a type of machine learning model that can be used to predict the spread of information or influence in a network. These models are often used in a variety of applications, such as social media marketing, viral prediction, and recommendation systems. One challenge in using diffusion models\cite{DBLP:journals/corr/Sohl-DicksteinW15}\cite{DBLP:journals/corr/abs-2006-11239}\cite{DBLP:journals/corr/abs-1907-05600} is the need for high-quality prompts, which are short pieces of text that are used to initiate the diffusion process. 

MTTN contains a wide range of prompts, covering a variety of topics and styles. The prompts have been carefully curated and annotated to ensure their quality and relevance. We evaluate the performance of different models and show that they are able to generate prompts.

We believe that MTTN will be a valuable resource for researchers working on prompt generation, as well as other text-2-text generation tasks.

\section{Dataset Details}
\subsection{Text Collection}
For building the dataset we had to establish a balance between prompts that were used in real scenario and then some structured text for balance. To set the context, the actual samples that were used for image generation had been collected. However, the prompts were disjointed and disorganized, so there had to be some standardized text backing this data. To address this issue,injection of text from standard datasets representing different types of contextual prompts: MS-COCO\cite{DBLP:journals/corr/LinMBHPRDZ14} (Microsoft Cognitive Services Text Collection), WiT\cite{srinivasan2021wit} (Visual and Linguistic Experimental Series), Flickr30k\cite{DBLP:journals/corr/PlummerWCCHL15}, and Conceptual Captions\cite{sharma2018conceptual} was done. This helped  set the overall balance without overloading the original content with perfectly organized text and was able to strike a balance between already present contextual knowledge graphs and unseen samples.

\subsection{Preprocessing}
The idea was to make MTTN flexible to the user's action. Even though prompt generation was the primary task, the dataset flexibility depended on how well the data is processed and sorted, kind of like having multiple buckets of text that can enhance general text-to-text generation task. 
The first step began with removing emojis and special characters, because as everyone is aware they can seriously hinder the performance of any LLM. Second important step was to make buckets of words that were gradually removed after each step like trickling through a filter. In total there included 5 steps after the initial removal, these steps or stages represent the elimination of different parts of speech at different levels. Each step had a certain parts of speech being removed from it, like verbs, adjectives, adverbs, etc, which resulted at the final stage or stage 5 which only consisted of nouns, in other words objects and subjects; name, place, animal \& things if considered at the very rudimentary strata.

\section{Analysis}
\subsection{Dataset Stats}
The initial amount of text samples representing true prompts were around 1.3M(million) but to add context variable number of samples from different standard datasets as mentioned before were collected. Just talking numbers game, then the total extracted data would have been, 200K samples from WIT\cite{srinivasan2021wit} dataset, 600K samples from MS COCO\cite{DBLP:journals/corr/LinMBHPRDZ14} dataset, 300K samples Conceptual Captions\cite{sharma2018conceptual}, 150K samples from Flickr30k\cite{DBLP:journals/corr/PlummerWCCHL15} dataset. Each having their own set of rules for extraction and basic cleaning done before being added to the final list. In total now there is around 2.4M text samples giving the base of MTTN dataset.
Further more for the ease of work and to analyse the contents or for faster finetuning we have also created several splits by randomly sampling the whole dataset. The splits are as follows: MTTN-100K, MTTN-250K, MTTN-500k \& MTTN-1M

\vspace{0.4cm}
\begin{figure}[h!]
\centering
\includegraphics[width=0.5\textwidth]{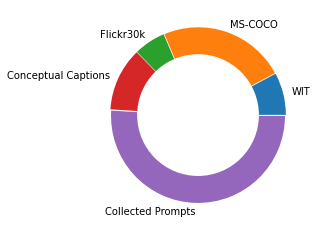}
\caption{Formation of MTTN Dataset}
\label{fig: MTTN dataset}
\end{figure}
\vspace{0.3cm}

\vspace{0.4cm}
\begin{figure}[h!]
\centering
\includegraphics[width=0.6\textwidth]{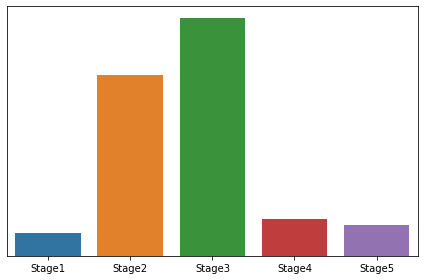}
\caption{Distribution of Words Removed at Each Stage}
\label{fig: Distribution of words removed at each stage}
\end{figure}
\vspace{0.3cm}

In MTTN we provide the original sample, in the first column followed by 5 more columns representing different trickling stages by reducing subsequent grammatical compositions. Each stage was formed  by masking words from the original prompt with the aim of regenerating the original prompts from the these stages . A list of words were masked during stage 1 and as the stages progressed  more words were masked subsequently. LLM models were sequentially fine-tuned over each stage to test how the models were adapting to the given data and how good the data was when considering the scenario of text generation. This also allowed us to reuse each model, from the previous stage on to the next rather than training it from scratch.
For analysis and base lining 3 different models were used, T5\cite{2020t5}, BART\cite{lewis2019bart}, and MVP\cite{tang2022mvp}. The models were each trained on the different stages sequentially as referred and the performances were captured for each of the stages. Below the results on the 100K split are drawn in a tabular format. Rouge score was used to draw the benchmark scores, for all the models.

\begin{center}
\begin{table}[h]
\vspace{0.4cm} 
\hspace*{3cm}\begin{tabular}{c c c c c c} 
\\
\hline \\[1ex]    
 Model Name    & Loss     & Rouge1   & Rouge2 & Rougel & Rougelsum \\ [1ex] 
 \hline \\[0.2ex]  
 BART & 0.1948 & 93.7086 & 86.409 & 93.4109 & 93.4199 \\ 
 [1ex] 

 T5 & 0.2611 & 93.3203 & 86.0067 & 93.0009 & 93.0012\\
 [1ex] 

 MVP & 0.2794 & 93.8372 & 87.5912 & 93.5366 & 93.5503\\
 [1ex] 
 \hline

\end{tabular}
\newline 
 \caption{Table 1: Results from Stage 1}
 \end{table}
\end{center}

\begin{center}
\begin{table}[h]
\vspace{0.4cm} 
\hspace*{3cm}\begin{tabular}{c c c c c c} 
\\
\hline \\[1ex]  
 Model Name    & Loss     & Rouge1   & Rouge2 & Rougel & Rougelsum \\ [1ex] 
 \hline \\[0.2ex]
 BART & 0.2616 & 93.3932 & 85.8573 & 93.105 & 93.1084 \\ 
 [1ex] 

 T5 & 0.2803 & 93.1422 & 85.4949 & 92.8251 & 92.818\\
 [1ex] 

 MVP & 0.3354 & 93.7784 & 87.2209 & 93.4807 & 93.4904\\
 [1ex] 
 \hline
\end{tabular}
\caption{Table 2 : Results from Stage 2}
 \end{table}
\end{center}

\begin{center}
\begin{table}[h]
\vspace{0.4cm} 
\hspace*{3cm}\begin{tabular}{c c c c c c} 
\\
\hline \\[1ex]  
 Model Name    & Loss     & Rouge1   & Rouge2 & Rougel & Rougelsum \\ [1ex] 
 \hline \\[0.2ex]
 BART & 0.5968 & 83.5774 & 65.4732 & 81.9868 & 81.9919 \\ 
 [1ex] 

 T5 & 0.6212 & 82.5331 & 63.1319 & 80.7276 & 80.7476\\
 [1ex] 

 MVP & 0.6400 & 84.344 & 67.7149 & 82.67 & 82.6912\\
 [1ex] 
 \hline
\end{tabular}
\caption{Table 3 : Results from Stage 3}
 \end{table}
\end{center}

\begin{center}
\begin{table}[h]
\vspace{0.4cm} 
\hspace*{3cm}\begin{tabular}{c c c c c c} 
\\
\hline \\[1ex]    
 Model Name    & Loss     & Rouge1   & Rouge2 & Rougel & Rougelsum \\ [1ex] 
 \hline \\[0.2ex]
 BART & 0.5938 & 82.5317 & 64.3066 & 80.6903 & 80.7076\\ 
 [1ex] 

 T5 & 0.6390 & 80.6725 & 60.797 & 78.7453 & 78.7481 \\
 [1ex] 

 MVP & 0.6015 & 83.6093 & 67.0664 & 81.9268 & 81.9398\\
 [1ex] 
 \hline
\end{tabular}
\caption{Table 4 : Results from Stage 4}
 \end{table}
\end{center}

\begin{center}
\begin{table}[h!]
\vspace{0.4cm} 
\hspace*{3cm}\begin{tabular}{c c c c c c} 
\\
\hline \\[1ex]  
 Model Name    & Loss     & Rouge1   & Rouge2 & Rougel & Rougelsum \\ [1ex] 
 \hline \\[0.2ex]
 BART & 0.5979 & 80.7564 & 62.2041 & 78.9507 & 78.9505\\ 
 [1ex] 

 T5 & 0.6718 & 78.7081 & 58.2488 & 76.5659 & 76.5574\\
 [1ex] 

 MVP & 0.6113 & 81.4961 & 64.5217 & 79.7526 & 79.75 \\
 [1ex] 
 \hline
\end{tabular}
\caption{Table 5 : Results from Stage 5}
 \end{table}
\end{center}

\newpage

\section{Future Work}

The entirety of MTTN is considered to be a very minuscule contribution to the world of NLP. We believe that MTTN has the potential to serve as a valuable foundation for a variety of natural language processing tasks. Its large volume and the wide range of combinations that can be created from it makes it a valuable resource. In the future, we would welcome any updates or additions to MTTN and hope to see it used in a variety of text generation and formation tasks.

\section{Conclusion}
The widespread traction of diffusion models\cite{DBLP:journals/corr/abs-2006-11239}\cite{DBLP:journals/corr/Sohl-DicksteinW15}\cite{DBLP:journals/corr/abs-1907-05600} has opened the way for further advancements in different areas combining NLP and CV. With this in mind we are releasing MTTN and its subsequent smaller splits. MTTN will pave the wave for more intuitive and complex text/prompt generation task with the help of different LLMs. We are also looking forward to different innovations that would be possible through the usage of MTTN. The volume of MTTN also allows for the creation of robust models that can generate coherent and coherent texts on a wide range of topics.

Finally contributors are always welcome, who can enhance this work and enrich MTTN, or create a derivative from this and pave the way for broader tasks.

To access MTTN, please follow this \underline{\textbf{\href{https://github.com/mttn2023/mttn}{Github Page}} }

\section*{Acknowledgments}
This work is supported by the all the contributors to different open source datasets, along with all the people who are contributing to different open-source platforms and libraries without which our work would have been incomplete.
Finally a huge thanks to all the people at huggingface\cite{wolf-etal-2020-transformers} for creating a platform rather a playground of ideas.
\\

\bibliographystyle{unsrt}  
\bibliography{references}

\end{document}